\documentclass{article}
\usepackage{amsfonts}
\usepackage{spconf,amsmath,graphicx,epsfig}
\usepackage{subcaption}
\usepackage{mwe}
\usepackage{xcolor}
\usepackage{hyperref}
\usepackage{algorithm}
\usepackage{algorithmic}
\usepackage{multirow}
\title{Improving accuracy of rare words for RNN-Transducer through unigram shallow fusion}
\name{\begin{tabular}{c} Vijay Ravi$^{1,2, \star}$ \qquad  Yile Gu$^{1\star}$ \qquad Ankur Gandhe$^{1}$ \qquad Ariya Rastrow$^{1}$ \qquad  \\ Linda Liu$^{1}$ \qquad Denis Filimonov$^{1}$ \qquad  Scott Novotney$^{1}$ \qquad Ivan Bulyko$^{1}$\thanks{$^\star$equal contribution}\thanks{$^2$work done as an applied scientist intern in Amazon Alexa}\end{tabular}}

\address{$^1$Amazon Alexa, USA \\$^2$Dept. of Electrical and Computer Engineering, University of California at Los Angeles, USA}

\begin{document}
\ninept
\maketitle
%  with DNN-based speech enhancement methods
\begin{abstract}
End-to-end automatic speech recognition (ASR) systems, such as recurrent neural network transducer (RNN-T), have become popular, but rare word remains a challenge. In this paper, we propose a simple, yet effective method called unigram shallow fusion (USF) to improve rare words for RNN-T. In USF, we extract rare words from RNN-T training data based on unigram count, and apply a fixed reward when the word is encountered during decoding. We show that this simple method can improve performance on rare words by 3.7\% WER relative without degradation on general test set, and the improvement from USF is additive to any additional language model based rescoring. Then, we show that the same USF does not work on conventional hybrid system. Finally, we reason that USF works by fixing errors in probability estimates of words due to Viterbi search used during decoding with subword-based RNN-T.

\end{abstract}

\section{Introduction}
\label{sec1}
%- speech recognition task
%- rise of e2e ASR
%- struggle with rare words
%- external knowledge (lexicon)
%- additional available knowledge (semantic labels)
%- the rest of this paper...

End-to-end models for automatic speech recognition (ASR) have gained popularity in recent years due to competitive performance and architectural simplicity, as they fold separate components of a conventional ASR system (i.e. acoustic, pronunciation, and language models (LM)) into a single neural network. Among these end-to-end models, recurrent neural network transducer (RNN-T)~\cite{graves2012sequence} is the most suitable as a streaming end-to-end recognizer, which studies~\cite{rao2017exploring,he2019streaming} have shown to achieve competitive performance compared to conventional systems. 

Despite competitive performance, RNN-T models lag behind on rare words accuracy compared to a conventional hybrid ASR system (such as \cite{povey11kaldi}). Transcribing those rare words is often critical for understanding the meaning of the utterance, as they are likely to be named entities such as names and locations. 

As RNN-T models are trained with audio-text pairs, to improve accuracies, the previous study~\cite{he2019streaming} applies a text-to-speech (TTS) system to text-only data and train an end-to-end model on the resulting audio as an alternative way to ingest text-only data. Conversely, to ingest additional audio-only data, the previous study\cite{li2019semi} applies a conventional model to decode audio-only data that contain proper nouns and train an end-to-end model on the transcripts. 

To integrate additional text-only data, ``LM fusion'' methods have been proposed. LM and RNN-T logits are interpolated during inference in shallow fusion~\cite{gulcehre2015using}, while LM is incorporated into neural architecture of end-to-end models in more sophisticated methods such as deep fusion~\cite{gulcehre2015using} and cold fusion~\cite{sriram2017cold}. In particular, a recent work~\cite{peyser2020improving} applies shallow fusion to incorporate a very large text corpus into RNN-T model to improve tail performance, and shows the advantage of intelligent pruning of text corpus and incorporation of LM in minimum word error rate (MWER) finetuning of RNN-T model. 

Several second pass rescoring methods have also been studied for improving RNN-T models. In addition to rescoring with regular neural LM~\cite{li2020developing,guo2020efficient},  a Listen, Attend, and Spell (LAS) component~\cite{li2020towards, sainath2019two} (which attends to acoustics) and recently a deliberation model~\cite{hu2020deliberation} (which attends to both acoustics and first-pass hypotheses) have been used. 
A recent study ~\cite{peyser2020improvingmwer} obtains further gains by modifying MWER loss criteria for the two-pass framework with LAS targeted at improving recognition of proper noun. 

This paper proposes a new, simple yet effective way to improve rare words accuracy for RNN-T models via unigram shallow fusion (USF). Unlike previous models, the proposed model has very small additional memory footprint (a few megabytes), and essentially adds no latency. In addition, because of its simplicity, it does not require training and thus is extremely easy to maintain and update. Furthermore, this paper will also show that the improvement from unigram shallow fusion is additive to the improvement to second pass rescoring. All these qualities make it an desirable choice for improving rare words accuracy for RNN-T in a production environment. 

The rest of this paper is organized as follows: we describe details of the proposed model and experimental setup in in Section~\ref{exp},  and present results in Section~\ref{res}. Finally, we conclude in Section~\ref{con} and outline future work.

\section{Experiments}
\label{exp}

\subsection{Unigram Shallow Fusion (USF)}
\label{section_usf}

\begin{figure}[ht!]
\begin{center}
\vspace{-2.4mm}
   \includegraphics[width=0.15\linewidth]{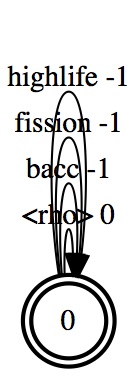}
\end{center} 
\vspace{-0.2cm}
   \caption{An illustration of FST used for unigram shallow fusion (USF). Here, rare words ``highlife'', ``fission'', and ``bacc'' are included in USF weights of $-1$ each. An additional ``otherwise'' arc $\rho$ is added with weight of $0$.  
   } 
\label{figure}
\end{figure}

Figure~\ref{figure} illustrates the unigram Finite-State Transducer (FST) that we use in shallow fusion (USF), where we include three words in USF. For each word included, an arc with score of $-1$ is added; an additional faillure arc $\rho$ (an ``otherwise'' arc) is added with weight of $0$ . To select words to be included in USF, we extract unigram counts from the human transcriptions used as RNN-T training data. The total total vocabulary size is 442k. We remove singletons to prevent spelling errors, and the vocabulary size becomes 220k. We then select words to be included in USF whose counts are between 2 and $n_\mathrm{thresh}$. The size of the unigram FST is small (i.e., $\sim$1 Megabytes for  $\sim$200k words), allowing it to have a small memory footprint.

 The mechanism of unigram shallow fusion (USF) applied in this paper is very similar to the previous work of contextual biasing for both hybrid~\cite{hall2015composition} and end-to-end~\cite{williams2018contextual,he2019streaming} ASR systems. However, previous works only focus on boosting scores for a small collection of n-grams based on contextual signals, e.g. dialog states~\cite{williams2018contextual}, domains~\cite{zhao2019shallow}, or personalization~\cite{huang2020class}. Unlike these studies, in this work, we show that we can perform shallow fusion with a simple unigram composed of a large collection of words ($\sim$200k), and we can activate all the time without contextual signals. By this simple unigram shallow fusion, we found that we can significantly reduce improve recognition on rare words, while not impact general test set.
\subsection{RNN-T and Neural Language Model (NLM)}

RNN-T score $\log P(y \mid x)$ (subword sequence $y$ given audio $x$) s then on-the-fly rescored by this score ($\log P_{USF}(y)$) queried from this unigram FST. As shown in Equation~\eqref{interpolation_weights} below, the combined score is scaled by the number of words ($|w|$) in the hypothesis, and then rescored by a NLM ($\beta \log P_{NLM}(y)$) and a coverage term ($ \gamma |w|$) through second pass rescoring. $ \gamma |w|$ is a token insertion term to promote longer transcripts as typically done for sequence-to-sequence models~\cite{chorowski2016towards}
During inference, our objective is to find the most likely subword sequence given the score from RNN-T model, USF, and NLM. In the equation, \emph{otf} and \emph{sp} indicate on-the-fly and second pass rescoring correspondingly. 

\begin{align*}
\hat{y}=\arg \max_{y} \{(\log P(y \mid x)+[\alpha \log P_{USF}(y)]_\mathrm{otf})/|w| + \\
 [\beta \log P_{NLM}(y) + \gamma |w|]_\mathrm{sp}\}
 \stepcounter{equation}\tag{\theequation}\label{interpolation_weights} 
\end{align*}

Both word level and subword level (similar to studies \cite{he2019streaming,zhao2019shallow}) shallow fusion are explored. 
For the majority of results reported below, we focus on word level shallow fusion. We study the impact of using subword-level in Table~\ref{table:subword}.

Similar to~\cite{guo2020efficient}, the RNN-T encoder includes of five LSTM layers. Each layer has 1024 hidden units. The prediction network of RNN-T has two LSTM layers of 1024 units and an embedding layer of 512 units. The softmax layer consists of 4k subword output units, and unigram language model~\cite{kudo2018subword} is used for subword segmentation. The NLM is word based, and contains a word embedding matrix of dimension 512 and two LSTM
layers of dimension 512 hidden units~\cite{raju2019scalable}.

\subsection{Datasets}

Both RNN-T and NLM are trained with $\sim$23,000 hour training set consisting of English utterances for voice control. We also augment the acoustic training data with the SpecAug algorithm \cite{park2019specaugment} to improve RNN-T model. 
For evaluation, we use two evaluation sets: (1) a test set $D_\mathrm{gen}$ of $149$ hours that represents the general use case (2) a rare words test set $D_\mathrm{rare}$ of $46$ hours. Starting from a large general use case test set, $D_\mathrm{rare}$ only includes utterances which contains at least a word that appears less than four times in the test set. All corpora contain non-identifiable utterances.

\section{Results}
\label{res}

\subsection{Improvements on rare words}

In Table~\ref{table:first}, we apply USF with $\alpha=0.75$ and $n_\mathrm{thresh}=250$ to RNN-T without and with second pass NLM activated. We report both WER relative (WERR) and oracle WERR. The oracle WER is computed based on selecting the hypothesis in the 8-best that minimizes WER for each utterance.  Comparing first two rows, one can see that USF improves rare words test $D_\mathrm{rare}$ by 3.8\% WERR, with a small degradation (-0.3\% WERR) on general test set $D_\mathrm{gen}$. Between next two rows where NLM is activated, one can see that the addition of USF still improves rare words test set. Crucially, the improvement from second pass NLM and USF are found to be additive. Furthermore, the presence of NLM alleviates the degradation from USF on $D_\mathrm{gen}$ (from -0.3\% to -0.1\%) which is consistent with a smaller degradation in oracle WER on $D_\mathrm{gen}$ from USF. Table~\ref{table:example} demonstrates examples where the addition of USF is able to correct recognition results.

\begin{table}[ht!]
 \centering
\begin{tabular}{|l|c|c|}
\hline
Model                    &$D_\mathrm{rare}$                     &$D_\mathrm{gen}$\\ \hline
RNN-T & baseline                   &     baseline                                   \\
RNN-T+ USF    & 3.8\% (1.0\%)                 & -0.3\%     (-0.1\%)                         \\ 
RNN-T+ NLM    & 0.9\%                         & 0.5\%                                      \\
RNN-T+ NLM+ USF                       & 4.6\%                &   0.4\%         \\ \hline
%Residual + RWMA                          & \textbf{0.982}  & \textbf{ 0.902}                  & \textbf{1.02}                \\ \hline
\end{tabular}
\caption{WER relative for different USF interpolation weights on rare words test set ($D_\mathrm{rare}$) and general test set ($D_\mathrm{gen}$). Oracle WER relative is included in parentheses. Baseline WER and baseline oracle WER for $D_\mathrm{rare}$ are around 2 times and 3 times of those for $D_\mathrm{gen}$, respectively. Positive means improvement. USF is applied with $\alpha=0.75$ and $n_\mathrm{thresh}=250$.}
\label{table:first}
\end{table}

\begin{table}[ht!]
  \centering
\begin{tabular}{|l|l|}
\hline
Models & Transcription \\ \hline 
w/o USF &  is \textbf{fishing} the opposite of fusion \\
w/ USF & is \textbf{fission} the opposite of fusion \\ \hline
w/o USF &  is anyways \textbf{in proper} english \\
w/ USF & is anyways \textbf{improper} english \\ \hline
w/o USF & play \textbf{back} at it again \\
w/ USF  & play \textbf{bacc} at it again \\ \hline
w/o USF & play ghana \textbf{hi life} in africa \\
w/ USF &  play ghana \textbf{highlife} in africa \\ \hline
w/o USF & search up \textbf{a and d y} on youtube \\
w/ USF  & search up \textbf{andy detwiler} on youtube \\ \hline
\end{tabular}
\caption{Examples of utterances that are corrected when USF is added.}
\label{table:example}
\end{table}

Due to the complementary effects between NLM and USF as discussed before, we apply NLM in all of the following results, and use NLM without USF as a baseline.

\subsection{Effects of interpolation weight $\alpha$ }

In Table~\ref{table:usf_level}, we explore the effects of interpolation weight of USF $\alpha$ from Equation~\eqref{interpolation_weights} on WER. One can see that increasing value for $\alpha$ at first further improves WERR on $D_\mathrm{rare}$, before the gain is diminished. For $D_\mathrm{gen}$, increasing $\alpha$ exacerbates the degradation. $\alpha$ of 0.75 strikes a good balance between improving $D_\mathrm{rare}$  while keeping $D_\mathrm{gen}$ not degraded.

\begin{table}[ht!]
\centering
\begin{tabular}{|l|c|c|}
\hline
$\alpha$                  &$D_\mathrm{rare}$       &$D_\mathrm{gen}$\\ \hline
w/o USF                &     baseline             & baseline                      \\
0.50                &     2.9\%   (0.8\%)         & 0.0\%  (0.0\%)                    \\
0.75   & 3.7\%   (1.0\%)           & -0.1\%              (-0.1\%)      \\ 
1.0   &4.2\%      (1.0\%)                & -0.5\%             (-0.1\%)                        \\
2.0       &2.3\%   (0.2\%)            &  -1.4\%    (-0.7\%)     \\ \hline
%Residual + RWMA                          & \textbf{0.982}  & \textbf{ 0.902}                  & \textbf{1.02}                \\ \hline
\end{tabular}
\caption{WER relative for different USF interpolation weights ($\alpha$) on rare words test set ($D_\mathrm{rare}$) and general test set ($D_\mathrm{gen}$). Oracle WER relative is included in parentheses. Positive means improvement. Second pass rescoring with NLM is activated for all cases. USF is applied with $n_\mathrm{thresh}=250$.}
\label{table:usf_level}
\end{table}

\subsection{Effects of number of unigrams}

As discussed in Section~\ref{section_usf}, words between 2 and $n_\mathrm{thresh}$ in unigram count in RNN-T training data are included in USF. In Table~\ref{table:usf_threshold}, we vary $n_\mathrm{thresh}$ and study its impact. The second column displays the number of words included in USF based on $n_\mathrm{thresh}$. One can see that increasing $n_\mathrm{thresh}$ (more words are included in USF) initially increases and then reduces WERR gain on $D_\mathrm{rare}$. When all the words whose counts above 1 are included, we see a degradation in WER (-16.0\%). There is a large jump in WERR between $n_\mathrm{thresh}$ of 31 and 250, which suggest that USF with those 30k words account for majority of the WERR from 3.7\% WERR in $n_\mathrm{thresh}$ of 250.

\begin{table}[ht!]
\centering
\begin{tabular}{|l|c|c|c|}
\hline
$n_\mathrm{thresh}$             &unigram \#      &$D_\mathrm{rare}$       &$D_\mathrm{gen}$\\ \hline
w/o USF &  -                   &     baseline             & baseline                      \\
8 & 137k                      & -0.1\%  (0.0\%)             &  -0.1\%   (-0.2\%)     \\ 
31 & 176k       &0.7\% (0.3\%)               &  -0.2\%  (-0.2\%       \\ 
250 &204k  &3.7\%    (1.0\%)                     & -0.1\%       (-0.1\%)                              \\
500 &209k  &4.0\%    (1.0\%)                     & -0.4\%       (-0.1\%)                              \\
all  &220k  & -16.0\% (-7.7\%)             & -12.7\%     (-9.0\%)               \\  \hline
%Residual + RWMA                          & \textbf{0.982}  & \textbf{ 0.902}                  & \textbf{1.02}                \\ \hline
\end{tabular}
\caption{WER relative on rare words test set ($D_\mathrm{rare}$) and general test set ($D_\mathrm{gen}$), for USFs with different thresholds for selecting words to be included in USF. Specifically, words whose counts in RNN-T training data is between 2 and $n_\mathrm{thresh}$ are included in USF. In the last row, all words whose counts are above 1 are included. The second column displays the number of words included in USF. Oracle WER relative is included in parentheses, and positive means improvement. Second pass rescoring with NLM is activated for all cases. USF is applied with $\alpha=0.75$.}
\label{table:usf_threshold}
\end{table}

\subsection{Effects of subword vs word level USF}

In previous studies on contexual biasing for RNN-T models, it is found (e.g.~\cite{he2019streaming,zhao2019shallow}) that subword level shallow fusion yields better results than word level shallow fusion. As a result, we explore results when subword level USF is applied in Table~\ref{table:subword}. Comparing Table~\ref{table:usf_level} and Table~\ref{table:subword} where same conditions are applied except for subword vs word level USF, one can see that subword level peforms much worse than word level.

\begin{table}[ht!]
\centering
\begin{tabular}{|l|c|c|}
\hline
$\alpha$                  &$D_\mathrm{rare}$       &$D_\mathrm{gen}$\\ \hline
w/o USF                &     baseline             & baseline                      \\
0.50                &     1.9\%   (0.7\%)         & -0.5\%  (-0.1\%)                    \\
0.75   & 1.8\%   (0.6\%)           & -1.5\%              (-0.7\%)      \\ 
1.0       &1.0\%   (0.3\%)            &  -4.2\%    (-1.6\%)     \\ \hline
%Residual + RWMA                          & \textbf{0.982}  & \textbf{ 0.902}                  & \textbf{1.02}                \\ \hline
\end{tabular}
\caption{WER relative when \emph{subword USF} is applied (with varying $\alpha$) on rare words test set ($D_\mathrm{rare}$) and general test set ($D_\mathrm{gen}$). Oracle WER relative is included in parentheses. Positive means improvement. Second pass rescoring with NLM is activated for all cases. USF is applied with $n_\mathrm{thresh}=250$.}
\label{table:subword}
\end{table}

This result reinforces the idea that the mechanism through which shallow fusion improves WER is different between the present study and previous works on contextual biasing. In contextual biasing, a contextual signal is present and we have a list of n-grams associated with it, which we want RNN-T to bias towards. While word level shallow fusion can only boost words after the words have been generated, subword level shallow fusion allows boosting part of the word before it is generated. Thus, subword level boosting allows RNN-T to have more chances to bias towards those words associated with the contextual signal. In the present study, however, we don't have contextual signals, and we are boosting much larger number of words. Furthermore, we apply unigram, as opposed to larger grams in contextual biasing. Changing from word to subword level shallow fusion would thus boost a huge list of subwords, which leads to lower precision in shallow fusion. Instead of biasing RNN-T to as in contextual biasing, what USF seems to do is to correct RNN-T scores for rare words from RNN-T -- we would discuss more about this in Section\ref{why_usf_works}.

\subsection{USF for hybrid model}

We also apply unigram shallow fusion for a hybrid ASR model to see whether we can see similar improvements. For hybrid model, the acoustic model is trained with same data as the RNN-T model, and it is a low-frame-rate model with 2-layer frequency LSTM \cite{li2017acoustic} followed by a 5-layer time LSTM trained with connectionist temporal classification (CTC) loss \cite{graves2006connectionist}.  The first-pass LM is a Kneser-Ney~\cite{kneser1995improved} smoothed n-gram LM, and its score is interpolated with second pass NLM which has the same structure as the one used for RNN-T rescoring. Both LMs are trained with the same data as the RNN-T model, with an additional small weight for automatic transcripts and external data sources.

The results for hybrid ASR system are shown in Table~\ref{table:subword}. We use results from hybrid model without USF as the baseline. We then apply USF as on-the-fly rescoring with varying $\alpha_\mathrm{hybrid}$, the interpolation weight applied to USF when combining score with hybrid model. Increasing $\alpha_\mathrm{hybrid}$ further improves WERR on $D_\mathrm{rare}$ for the range of values considered. However, unlike RNN-T system as in Table~\ref{table:usf_level}, the degradation in $D_\mathrm{gen}$ is severe even with a small $\alpha_\mathrm{hybrid}$. To have a minimal degradation on $D_\mathrm{gen}$, only 0.6\% WERR can be achieved for the hybrid model on $D_\mathrm{gen}$, unlike 3.7\% as for RNN-T. 

\begin{table}[ht!]
\centering
\begin{tabular}{|l|c|c|}
\hline
$\alpha_\mathrm{hybrid}$                  &$D_\mathrm{rare}$       &$D_\mathrm{gen}$\\ \hline
w/o USF                &     baseline             & baseline                      \\
0.25                &     0.6\%   (0.5\%)         & -0.1\%  (-0.2\%)                    \\
0.50                &     1.1\%   (1.1\%)         & -0.3\%  (-0.4\%)                    \\
0.75   & 1.6\%   (1.6\%)           & -0.6\%              (-0.6\%)      \\ 
1.5   & 2.4\%   (2.5\%)           & -1.8\%              (-1.3\%)      \\ 
2.5       &2.6\%   (2.2\%)            &  -4.6\%    (-3.6\%)     \\ \hline
%Residual + RWMA                          & \textbf{0.982}  & \textbf{ 0.902}                  & \textbf{1.02}                \\ \hline
\end{tabular}
\caption{WER relative for \emph{hybrid ASR system} with varying $\alpha$ on rare words test set ($D_\mathrm{rare}$) and general test set ($D_\mathrm{gen}$). Oracle WER relative is included in parentheses. Positive means improvement.  USF is applied with $n_\mathrm{thresh}=250$.}
\label{table:subword}
\end{table}

\subsection{Why USF works}
\label{why_usf_works}
The observation that USF works well on RNN-T but not hybrid system suggests that USF is to fix some inherent bias with RNN-T model. Since RNN-T models are trained typically with subword units~\cite{rao2017exploring} but the final output is words, we can get the probability of a word $w$ given a sequence of audio $x$ as: 
\begin{equation}
\label{eq:sum}
P(w|x) = \sum_{\{seg_{w}\}} P(w|seg_w)* P(seg_w|x) \\
= \sum_{\{seg_{w}\}} p(seg_w|x)
\end{equation} 
where $\{seg_{w}\}$ is the set of all possible segmentations of the word $w$. 

However, during actual inference, Viterbi (max) decoding is used (to be compute efficient) to get the best possible sequence of words for a given audio~\cite{graves2012sequence} and only the maximum probability $P(seg_w|x)$ is considered to compute the final likelihood of a word. The difference between this likelihood and the one in Equation~\eqref{eq:sum} results in \emph{search errors} (as opposed to \emph{model errors}). These differences (and thus the search errors) could become more severe for rare words which tend to be longer and have more ways of segmentation (on the other end, for frequent words, the subword is the word itself and thus there is no search error). USF proposed in this paper exploits shallow fusion to compensate such search errors for rare words -- without having to do the expensive sum (log semiring). As a followup work, we plan to work on verifying this empirically, and use it to assign more optimal scores in USF to maximize gains. 

\section{Conclusions}
\label{con} 
In this paper, to improve rare words for RNN-T, we proposed a simple, yet effective method unigram shallow fusion (USF). We showed that this method can improve performance on rare words by 3.7\% WERR without degradation on general test set, and the improvement from USF is additive to second pass rescoring.  Finally, we reasoned that USF works by fixing errors in probability estimates of words due to Viterbi search used during decoding in subword-based RNN-T.

We hope that this study not only shares a practical method to improve rare words for RNN-T, but also can inspire future work on better training of RNN-T model itself, e.g. it would be interesting to see whether we can modify RNN-T training loss to account for unintended underestimation of probabilities of rare words as reported in this study.

\section{Acknowledgment}

The authors would like to thank Fan Chen, Richard Diehl Martinez, Harish Arsikere, and Lingnan Lu from Amazon for helping with data setup. 

%\clearpage
\bibliographystyle{IEEEbib}
\bibliography{refs}

\end{document}